\documentclass{article} 
\usepackage{iclr2026_conference,times}
\iclrfinalcopy

\usepackage{amsmath,amsfonts,bm}

\def\eqref#1{equation~\ref{#1}}

\def\1{\bm{1}}

\DeclareMathAlphabet{\mathsfit}{\encodingdefault}{\sfdefault}{m}{sl}
\SetMathAlphabet{\mathsfit}{bold}{\encodingdefault}{\sfdefault}{bx}{n}

\usepackage{graphicx}
\usepackage{hyperref}
\usepackage{wrapfig}
\usepackage{url}
\usepackage[ruled,vlined]{algorithm2e}
\usepackage{hyperref}
\hypersetup{colorlinks=true,linkcolor=red,citecolor=green}
\usepackage{url}

\title{Zero-Residual Concept Erasure via Progressive Alignment in Text-to-Image Model}

\author{
Hongxu Chen$^1$, Zhen Wang$^1$, Taoran Mei$^2$, Lin Li$^1$, Bowei Zhu$^2$, Runshi Li$^2$, Long Chen$^1\thanks{Long Chen is the corresponding author.}$ \\
$^1$The Hong Kong University of Science and Technology \\
$^2$University of Science and Technology of China \\
\texttt{\{hchenej, zwangjr\}@connect.ust.hk, \{lllidy, longchen\}@ust.hk} \\
\texttt{\{meitaoran,zbw\_ustc,stflrs\}@mail.ustc.edu.cn} \\
}

\newcommand{\ie}{\textit{i}.\textit{e}.}
\newcommand{\eg}{\textit{e}.\textit{g}.}
\newcommand{\cf}{\textit{cf.}}

\begin{document}

\maketitle
\begin{figure}[h]
\begin{center}
\includegraphics[width=1.0\linewidth]{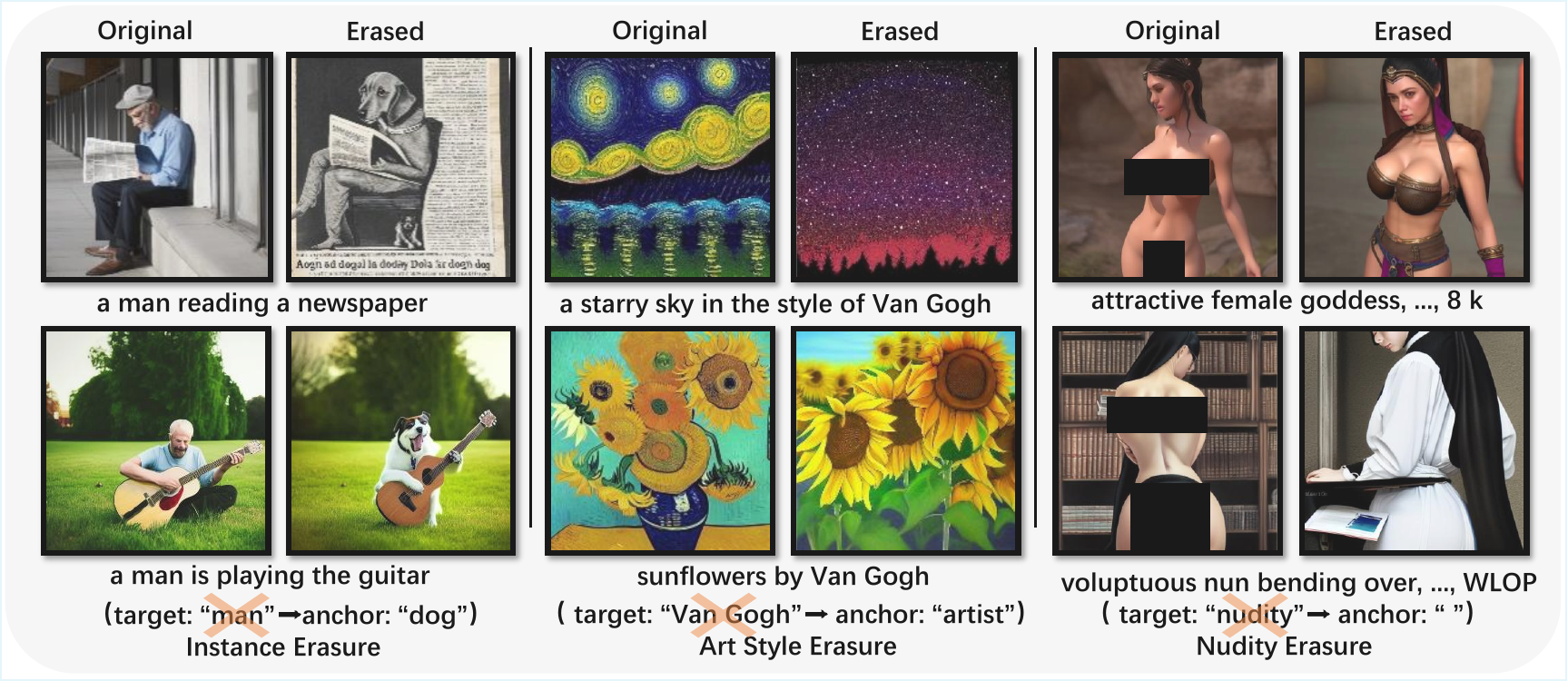}
\end{center}
\vspace{-1.5em}
\caption{Overview of the applications of ErasePro across multiple tasks in text-to-image model, including instance erasure (\eg, erasing ``\texttt{man}''), art style erasure (\eg, erasing ``\texttt{Van Gogh}''), and nudity erasure (\eg, erasing ``\texttt{nudity}''). Our method can remove undesirable target concepts by mapping them to anchor concepts which are semantically harmless or user-desired.}
\label{fig:pic-1}
\end{figure}

\begin{abstract}
Concept Erasure, which aims to prevent pretrained text-to-image models from generating content associated with semantic-harmful concepts (\ie, \emph{target concepts}), is getting increased attention. State-of-the-art methods formulate this task as an optimization problem: they align all target concepts with semantic-harmless \emph{anchor concepts}, and apply closed-form solutions to update the model accordingly. While these closed-form methods are efficient, we argue that existing methods have two overlooked limitations: \textbf{1)} They often result in incomplete erasure due to ``non-zero alignment residual'', especially when text prompts are relatively complex. \textbf{2)} They may suffer from generation quality degradation as they always concentrate parameter updates in a few deep layers.
To address these issues, we propose a novel closed-form method \textbf{ErasePro}: it is designed for more complete concept erasure and better preserving overall generative quality. Specifically, ErasePro first introduces a strict zero-residual constraint into the optimization objective, ensuring perfect alignment between target and anchor concept features and enabling more complete erasure. Secondly, it employs a progressive, layer-wise update strategy that gradually transfers target concept features to those of the anchor concept from shallow to deep layers. As the depth increases, the required parameter changes diminish, thereby reducing deviations in sensitive deep layers and preserving generative quality. Empirical results across different concept erasure tasks (including instance, art style, and nudity erasure) have demonstrated the effectiveness of our ErasePro. 
\end{abstract}

\section{Introduction}

Pretrained text-to-image (T2I) models (\eg, stable diffusion and DALL-E 2)~\cite{saharia2022photorealistic, rombach2022high, ramesh2022hierarchical, peebles2023scalable, chang2023muse, yu2022scaling} have shown impressive performance across diverse generative tasks. By utilizing customized prompts, users can easily steer these models to generate realistic and high-quality images. 
However, due to the presence of inappropriate content~\cite{shan2023glaze, somepalli2023diffusion, carlini2023extracting} in their noisy web-scraped training data~\cite{schuhmann2021laion, schuhmann2022laion}, 
these models can be misused to generate harmful outputs, leading to serious ethical and legal concerns~\cite{setty2023ai, maleve2024style}.
For instance, as shown in Figure~\ref{fig:pic-1}, given prompts containing a specific concept (\eg, prompt \texttt{Sunflowers by Van Gogh}, where ``\texttt{Van Gogh}'' is the target concept), pretrained T2I models may generate content involving copyrighted artwork or nudity. In response to these concerns, a task named \textbf{concept erasure} has been proposed to prevent the generation of such unsafe or undesired target concepts in pretrained T2I models.

The most straightforward concept erasure approach is to simply filter out all generated images that contain target concepts~\cite{rando2022red}, or modify the generation process during inference~\cite{schramowski2023safe, wang2025precise, yoon2024safree} by steering latent representation away from the target concept. However, since these post-training methods do not modify the model's pretrained parameters, they are often vulnerable to circumvention~\cite{mano2022tutorial} when model weights or source codes are accessible, leading to superficial and unstable erasure.

\begin{figure}[h]
\begin{center}
\includegraphics[width=1.0\linewidth]{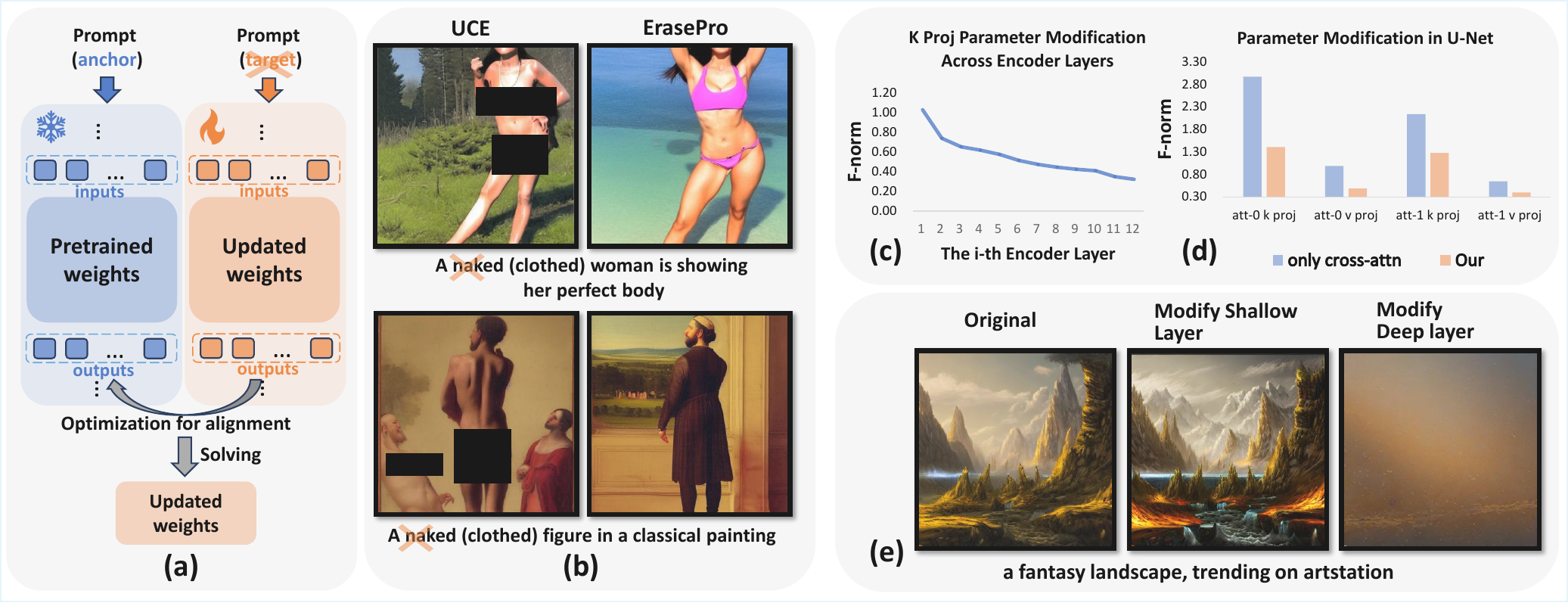}
\end{center}
\vspace{-1.5em}
\caption{\textbf{(a)} Prior closed-form approaches attempt to erase target concepts by solving an optimization problem that enforces alignment between target and anchor concept features, thereby substituting semantics of target concept with those of the anchor. \textbf{(b)} These methods suffer from incomplete erasure of target concepts. \textbf{(c)} When applying ErasePro for instance erasure in stable diffusion, our progressive framework ensures that modifications to the k-projection layers diminish with increasing depth of the text encoder, and \textbf{(d)} the update burden on the U-Net’s cross-attention (deep layers) is significantly reduced compared to directly modifying them. \textbf{(e)} We manually inject identity-based parameter deviations and compare their impact between layers of different depths. For the same deviation magnitude, deeper layers cause more severe degradation in generative quality\footref{footnote: appendix_setting}. }
\label{fig:pic-2}
\end{figure}

In contrast, more recent works focus on directly modifying model parameters to enforce the inherent forgetting of target concepts. Generally, these methods can be broadly categorized into two types: 1) \textbf{\emph{Gradient-based methods}}~\cite{kumari2023ablating, zhang2024forget, lu2024mace}: They employ specially designed loss functions to fine-tune pretrained models via gradient descent. Typically, by using prompts containing target concepts, they guide and fine-tune the model to suppress the presence of these target concepts in the generated images. Unfortunately, they often require a large number of curated prompt-image pairs and numerous gradient updates, making them computationally expensive. 2) \textbf{\emph{Closed-form methods}}~\cite{gandikota2024unified, li2025speed, gong2024reliable}: They formulate concept erasure as a parameter optimization problem. To mitigate catastrophic forgetting, they directly apply closed-form solutions to a few deep layers without gradient updates. Specifically, given prompts containing target concepts, their objective is to align the textual features of target concepts with semantically harmless \emph{anchor concepts} (\cf, Figure~\ref{fig:pic-2}(a)). By minimizing ``\emph{alignment residual}'', they enable effective erasure when provided prompts are relatively simple (\eg, a prompt only involves the single target concept, such as \texttt{naked figure}). Owing to the gradient-free nature, closed-form methods are significantly faster than gradient-based counterparts, while maintaining comparable performance. 

In this paper, we argue that almost all existing closed-form methods still have two overlooked limitations: 

\noindent{\textbf{``Incomplete'' Erasure:}} The alignment residual of previous optimization remains non-zero after applying their closed-form solution to the objective, \ie, they cannot perfectly align target concept features with anchor concept features in the optimization objective. Therefore, during the inference stage, given relatively complex prompts, the misalignment tends to amplify, leading to a higher risk of incomplete erasure. For example, as illustrated in Figure~\ref{fig:pic-2}(b), when attempting to map the concept of ``\texttt{naked}'' to ``\texttt{clothed},'' the existing method still generates photos with nudity when provided with relatively complex prompts. 

\noindent{\textbf{Generation Degradation:}} Since these methods typically update the parameters of a few deep layers (\eg, cross-attention module in U-Net~\cite{ronneberger2015u}), and these layers are highly correlated with the model’s generative capability~\cite{kumari2023multi,  staniszewski2025precise}, the ``update burden'' of these layers becomes heavy. This may result in large parameter deviations, leading to degradation of the model's overall generative quality. The problem is further exacerbated when there is a large semantic gap between the target and anchor concepts, resulting in greater parameter deviations and more severe degradation.

To overcome these limitations, in this paper, we propose a novel closed-form concept erasure method: \textbf{ErasePro}. Specifically, \textbf{1)} \emph{to address the issue of incomplete erasure}, we introduce a strict constraint into the objective. This new constraint guarantees that the alignment residual is exactly zero after applying the closed-form solution to the objective, enabling target concepts to perfectly align with anchor concepts. Consequently, it achieves a more complete erasure with complex prompts.
\textbf{2)} \emph{To preserve the model’s overall generative quality}, ErasePro adopts a progressive, layer-wise optimization framework. It updates the network from shallow to deep layers, enabling a gradual transition from target to anchor concepts. As this transition propagates, the required parameter deviations become increasingly subtle (\cf, Figure~\ref{fig:pic-2}(c)). This design shifts the update burden to the shallow layers and significantly reduces parameter deviations in deeper layers compared to directly updating them (\cf, Figure~\ref{fig:pic-2}(d)). 
Given that overall generative quality is more sensitive to changes in deeper layers than shallow ones\footnote{As evidenced by Figure~\ref{fig:pic-2}(e), equal-magnitude parameter deviations in deeper layers lead to greater degradation.}, our method can minimize deviations in deep layers to better preserve generative quality.

We evaluated our method on several concept erasure tasks, demonstrating its effectiveness. In summary, our \textbf{contributions} are three-fold: \textbf{1)} We propose ErasePro, a novel concept erasure algorithm, supporting various applications on instance, art style, and nudity erasure. \textbf{2)} We introduce a novel optimization objective and a progressive alignment framework to achieve complete concept erasure while better preserving overall generative quality. \textbf{3)} We empirically demonstrate the effectiveness of ErasePro, highlighting its significant improvements over state-of-the-art methods.

\section{Related Work} 

\noindent\textbf{Text-to-Image Generation} has made remarkable strides in recent years~\cite{saharia2022photorealistic, rombach2022high, ramesh2022hierarchical, peebles2023scalable, ramesh2022hierarchical, chang2023muse, yu2022scaling}, particularly with the emergence of diffusion-based models such as stable diffusion~\cite{rombach2022high}, DALL-E 2~\cite{ramesh2022hierarchical}, and DiT~\cite{peebles2023scalable}. Pretrained on large-scale datasets, these models demonstrate impressive generative capabilities, producing high-quality images which are semantically aligned with natural language prompts. To support personalized concept customization~\cite{gu2023mix, kumari2023multi, ruiz2023dreambooth, chen2025iteris}, various transfer learning techniques ~\cite{hu2022lora, li2021prefix, jia2022visual, liu2021p, diao2024unipt} techniques have also been widely adopted. 

\noindent\textbf{Gradient-based Concept Erasure}~\cite{kumari2023ablating, zhang2024forget, lu2024mace} suppresses target concepts by fine-tuning pretrained T2I models using gradient descent. For instance, AC~\cite{kumari2023ablating} employs prompts containing target concepts alongside hundreds of images representing anchor concepts to fine-tune the pretrained diffusion model. Specifically, it minimizes the KL divergence between the distributions of target and anchor concepts, thereby encouraging the latent representations of target concepts to align with those of the anchors.

\noindent\textbf{Closed-form Concept Erasure}~\cite{gandikota2024unified, li2025speed, gong2024reliable} approaches solve unconstrained optimization problems over a limited number of layers and update model parameters in a single step via closed-form solutions. A representative method is UCE~\cite{gandikota2024unified}, which modifies only the cross-attention layers of diffusion models. By aligning the textual features of target concepts with those of anchor concepts, UCE can effectively erase target concepts when provided with relatively simple prompts during inference. However, due to non-zero alignment residuals, it struggles to achieve complete erasure when dealing with more complex prompts. In contrast, our method is explicitly designed to enable a more complete concept erasure even for such complex cases.

\section{Methodology}
\subsection{Preliminaries}

\noindent\textbf{Problem Formulation.} 
Given a pretrained T2I model, concept erasure aims to suppress the generation of undesired target concepts. A common strategy introduces anchor concepts that serve as semantic substitutes for the target concepts during generation. To achieve this, we formulate optimization by aligning target concept features with anchor concept features, specifically within the layers that process only text features. Without loss of generality, we present the formulation in the context of a single representative layer with pretrained parameters $\mathbf{W_o}$.

Given a prompt (\eg, \texttt{naked figure}) with a concept (\eg, ``\texttt{naked}''), we perform inference in the model, extracting concept features by concatenating input sequences to this layer.
Given $N$ target-anchor concept pairs, a set of target concept features $\mathbf{X}=[\mathbf{x_1}, \dots, \mathbf{x_N}]$ and their corresponding anchor concept features $\mathbf{Y}=[\mathbf{y_1}, \dots, \mathbf{y_N}]$, our objective is to derive the layer’s parameter solution $\mathbf{W}$ such that each target feature $\mathbf{x_i}$ is mapped to the corresponding anchor feature $\mathbf{y_i}$, thereby achieving concept erasure.

\noindent\textbf{Closed-Form Formulation.}
State-of-the-art closed-form methods~\cite{gandikota2024unified, li2025speed, gong2024reliable} always modify the parameters of a few deep layers. In diffusion-based architectures, such modifications are typically applied to the cross-attention layer of the U-Net, which is responsible for integrating textual conditioning into the denoising process. These methods seek to alter the original parameters $\mathbf{W_o}$ of projection layer in the cross-attention, aiming for the target features $\mathbf{X}$ to align with the anchor features $\mathbf{Y}$ after the attention operation. This alignment residual is commonly measured by the Frobenius norm ($\|\mathbf{W X} - \mathbf{W_{o} Y}\|^2_{F}$). By encouraging this alignment, the model is implicitly guided to interpret the anchor concept as a semantic substitute for the target, thereby suppressing the generation of target concepts during inference. Formally, the objective is typically posed as the following unconstrained optimization problem or a close variant:
\begin{equation}
\small
\mathbf{W^{*}}=\arg\min_\mathbf{W} \left(\|\mathbf{W X} - \mathbf{W_{\text{o}} Y}\|^2_{F}+\|\mathbf{W-W_o}\|_{F}^2 \right).
\label{eq1}
\end{equation}
Here, $\mathbf{W}$ denotes the updated parameter matrix (\eg, the key-value projection layer in the cross-attention), and $\mathbf{W_{o}}$ represents the pretrained matrix. The primary term $\|\mathbf{WX}- \mathbf{W_{o} Y}\|^2_{F}$ measures the alignment residual between the transformed target and anchor features. In addition, the regularization term $\|\mathbf{W} - \mathbf{W_{o}}\|^2_{F}$ constrains the update magnitude, preventing $\mathbf{W}$ from deviating significantly from $\mathbf{W_{o}}$. 

This objective has the following closed-form solution:
\begin{equation}
\small
\mathbf{W^*} = \left( \mathbf{W_o YX^\top} + \mathbf{W_o} \right) \left( \mathbf{XX^\top} + \mathbf{I} \right)^{-1}.
\label{eq2}
\end{equation}

\begin{figure}[t]
    \centering
    \includegraphics[width=1.0\textwidth]{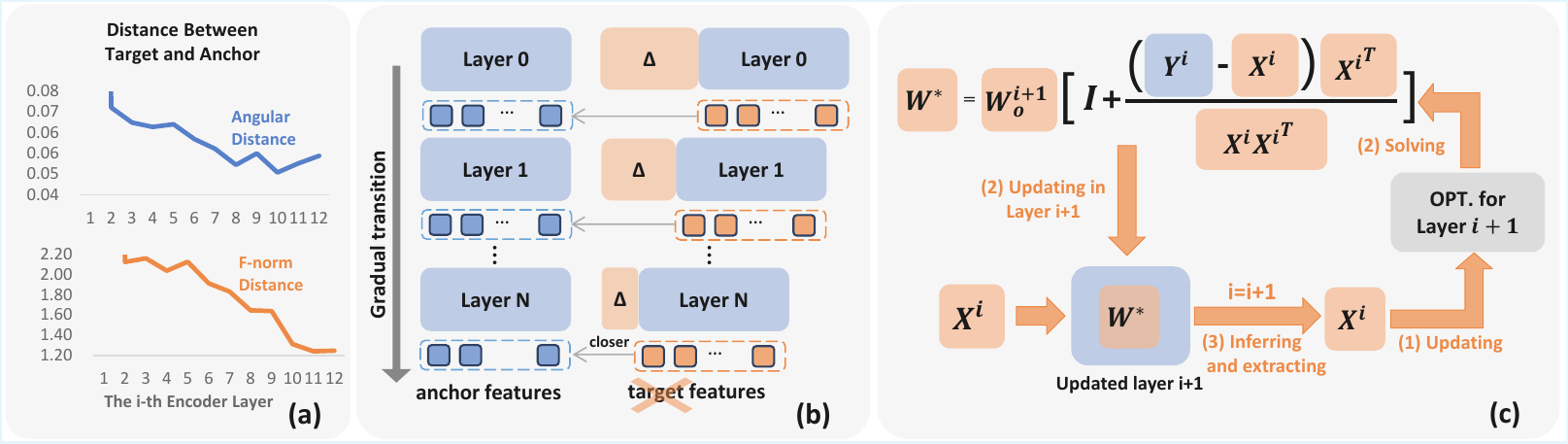}
    \vspace{-1em}
    \caption{\textbf{(a)} In our method, the target features progressively align with the anchor features, as indicated by the decreasing angular and Frobenius distance with network depth. \textbf{(b)} The magnitude of parameter deviations $\Delta$ gradually decreases in deeper layers as the target features progressively align with the anchor features. \textbf{(c)} Our method (ErasePro) employs a progressive alignment framework to effectively erase target concepts.}
    \label{fig:pic-4}
\end{figure}

\subsection{Proposed Approach: EarsePro}
 Unlike previous closed-form approaches that restrict updates to only a few layers, our method introduces a progressive alignment framework for multiple layers, driven by a new constrained formulation. This new formulation facilitates more complete alignment from target to anchor concepts, thereby improving the effectiveness of concept erasure. In addition, our optimization proceeds from shallow to deep, with the objective updated based on features extracted from the current model state. This gradual transition enables the target features to be smoothly aligned with the anchor features, while minimizing parameter deviations in deeper layers. Thus, the generative capacity is better preserved. 

In the following, we provide a detailed discussion of the two key improvements\footnote{The detailed derivation is left in the appendix.} introduced by our approach, along with empirical observations from the previous formulation.

\noindent\textbf{Observation 1:} \emph{Existing solutions can not minimize the alignment residual\footnote{We refer to the residual in the objective. From a full-module perspective, it is inherently non-zero due to residual connections.} to exactly zero in the objective.} The solution in Eq.~\ref{eq2} cannot minimize the primary objective (the alignment residual $\|\mathbf{WX}- \mathbf{W_{o} Y}\|^2_{F}$ in Eq.~\ref{eq1}) to exactly zero, indicating a rough alignment. In fact, substituting Eq.~\ref{eq2} into alignment residual yields the following:
\begin{equation}
\small
\left\|\mathbf{W_o} \left[(\mathbf{Y X^T} + \mathbf{I})(\mathbf{X X^T} + \mathbf{I})^{-1} \mathbf{X} - \mathbf{Y}\right] \right\|_F^2,
\label{eq5}
\end{equation}
In general, this alignment residual Eq.~\ref{eq5} remains non-zero in almost all cases, meaning that the target concept features cannot be perfectly aligned with those of the anchor concepts after updating. As a result, the model may still retain parts of the original target concepts.
This issue becomes more pronounced when utilizing semantically complex prompts. In such cases, the extracted features $\mathbf{X}$ may differ significantly from those used during optimization with simpler prompts. This mismatch leads to an even larger residual in alignment residual, further compromising the alignment quality between the target and anchor concepts. Consequently, target concepts may still appear in the generation.

\begin{algorithm}[!t]
\caption{Our algorithm (ErasePro)}
\KwIn{
    Prompt input $\mathbf{X}^0$, $\mathbf{Y}^0$; Layer number $S$;
    Pretrained weights across all layers $\{\mathbf{W_o^i}\}_{i=1}^S$
}
\KwOut{Updated model}

$\{\mathbf{Y^i}\}_{i=1}^{S} \leftarrow \text{Extract-Anchor-Features}(\text{Model}, \mathbf{Y^0})$\;

\For{$i = 1, \cdots, S$}{
    $\mathbf{W^*} \leftarrow \text{Solution}(\mathbf{Y^{i-1}}, \mathbf{X^{i-1}}, \mathbf{W_o^i})$; // Eq~\ref{eq4}\\
    $\text{Layer}^i \leftarrow \text{Update\_Weights}(\text{Layer}^i, \mathbf{W^*})$; 
    $\mathbf{X^i} \leftarrow \text{Layer}^i(\mathbf{X^{i-1}})$; \\
    $\text{Model} \leftarrow \text{Update}(\text{Model}, \text{Layer}^i)$\;
}
\Return Model\;
\label{algorithm}
\end{algorithm}

\noindent\textbf{Improvement 1:} \emph{ErasePro introduces a constrained formulation for zero-residual.}
To achieve alignment with zero residual error, our method directly enforces a hard constraint such that each $\mathbf{x_i}$ exactly aligns with its corresponding $\mathbf{y_i}$. Under this constraint, we further minimize the deviations between the updated and pretrained parameters. Formally, our objective is defined as follows:
\begin{equation}
\small
\mathbf{W^{*}}=\arg\min_\mathbf{W} \|\mathbf{W}-\mathbf{W_o}\|_{F}^2, s.t. \mathbf{W x_i} = \mathbf{W_{o} y_i},
\label{eq3}
\end{equation}
where $i=1, \dots, N$. The closed-form solution to this constrained optimization problem is:
\begin{equation}
\small
\mathbf{W^*} = \mathbf{W_o} + (\mathbf{W_o Y} - \mathbf{W_o X})(\mathbf{X ^\top X})^{-1}\mathbf{X^\top}.
\label{eq4}
\end{equation}
The constraint ensures the alignment residual $\|\mathbf{WX} - \mathbf{W_o Y}\|_F^2$ is zero after applying the solution in Eq.~\ref{eq4}, which means our method ensures a full alignment from the target features to the anchor features, thereby achieving more complete erasure for target concepts.

\noindent\textbf{{Observation 2:}} \emph{Existing solutions suffer from parameter deviations in deep layers, risking overall performance degradation.} 
Considering the closed-form solution obtained in Eq.~\ref{eq2}, its deviates from the pretrained parameters $\mathbf{W_o}$:
\begin{equation}
\Delta  = \mathbf{W^*} - \mathbf{W_o} = \mathbf{W_o} (\mathbf{Y} - \mathbf{X})\mathbf{X}^\top(\mathbf{X X^\top} + \mathbf{I})^{-1}.
\label{eq6}
\end{equation}

These methods typically update only a small subset of deep layers with limited parameters. In diffusion-based architectures, such updates are only applied to the cross-attention layers of the U-Net. When the update burden is restricted to these deep layers that are highly correlated with the model’s generative capability, these deviations can degrade the model’s overall generative quality. 
Moreover, when the semantic gap between the target and anchor concepts is large (i.e., when $\mathbf{Y}$ and $\mathbf{X}$ differ significantly), Eq.~\ref{eq6} implies that the magnitude of $\Delta $ increases. This suggests that parameter deviations become more substantial, potentially exacerbating generation degradation.

\noindent\textbf{Improvement 2}: \emph{ErasePro proposes a progressive alignment framework to shift the update burden on deep layers.}  Unlike prior closed-form approaches that modify only a few deep layers, our method expands the updating scope across multiple layers to reduce the update burden on these layers. 

Specifically, starting from the shallow layers (i.e., where textual prompts are initially processed) and progressing toward deeper layers (e.g., cross-attention), we sequentially apply closed-form updates to each layer.  As illustrated in Figure~\ref{fig:pic-4}(c), at the $i$-th stage targeting layer $i$, ErasePro performs the following steps: 
\textbf{1)} The input features $\mathbf{X}$, the anchor features $\mathbf{Y}$, and the pretrained parameters $\mathbf{W_{o}}$ are used to formulate a constrained optimization problem, as defined in Eq.~\ref{eq4};
\textbf{2)} Solving this optimization yields a closed-form solution $\mathbf{W}^{*}$, which is then used to update the parameters of layer $i$;
\textbf{3)} Inference is performed on the updated layer $i$ using the input features $\mathbf{X}$, extracting output features and updating $\mathbf{X}$, which are fed into the next-stage optimization.

As shown in Figure~\ref{fig:pic-4}(a), ErasePro progressively reduces the distance between target and anchor features, indicating this progressive alignment strategy allows the target features to be gradually aligned with the anchor features throughout the network. As illustrated in Figure~\ref{fig:pic-4}(b), since target features are already better aligned in deeper layers, the required parameter updates to achieve alignment diminish as the depth increases. In other words, our strategy effectively shifts the update burden to the shallow layers, which are less sensitive to overall generative quality, thereby reducing the risk of over-modifying more sensitive deep layers and preserving the model’s overall generative quality. In contrast, previous closed-form methods attempt to achieve alignment in a single step within restricted positions, indicating a heavy update burden in these deep layers. It is worth noting that this progressive framework also contributes to alleviating the incomplete erasure issue. As ErasePro performs multi-layer alignment, incrementally reducing the distance between target and anchor features at each alignment step, thereby further minimizing the alignment residual. The complete procedure of ErasePro is detailed in Algorithm~\ref{algorithm}.

\section{Experiments}
\subsection{Experimental Setting}

\begin{figure}[t]
    \centering
    \includegraphics[width=1.0\textwidth]{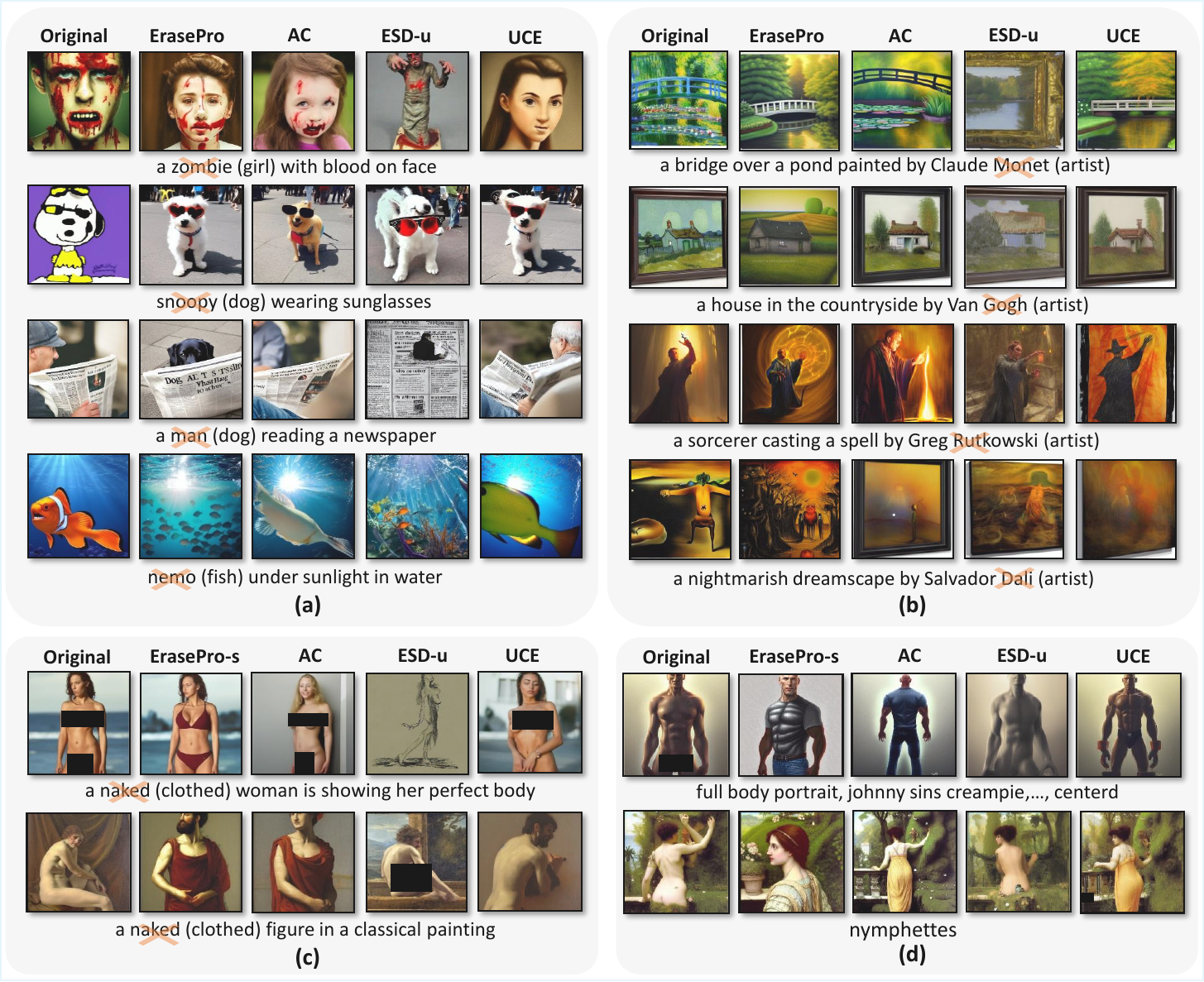}
    \vspace{-1em}
    \caption{\textbf{(a)} Qualitative results for instance erasure. \textbf{(b)} Qualitative results for art style erasure. \textbf{(c)} Qualitative results for explicit nudity erasure. \textbf{(d)} Qualitative results for implicit nudity erasure.}
    \label{fig:pic-4}
\end{figure}

\noindent\textbf{Baselines.} 
We compared ErasePro against five main baselines\footnote{The detailed setting is left in the appendix. \label{footnote: appendix_setting}}. Specifically, there are three gradient-based methods: \textbf{AC}~\cite{kumari2023ablating}, \textbf{ESD-x}, and \textbf{ESD-u}~\cite{gandikota2023erasing}. For closed-form methods, we adopt \textbf{UCE}~\cite{gandikota2024unified}. Pretrained model \textbf{SDv1.4}~\cite{rombach2022high} serves as the base model. It is worth noting that most existing methods are built upon UCE's design, \ie, share similar optimization objectives, and update the same components.
For nudity erasure, we also include \textbf{SDv2.1} as a baseline.

\noindent\textbf{Evaluation Setting and Metrics.} We first followed the general evaluation setting~\cite{kumari2023ablating} for all the erasure tasks, where the target concepts ``explicitly'' appear in the prompt (\eg, ``\texttt{naked figure}'' for the concept ``\texttt{naked}"). Specifically, for the erased model, we generated 200 images for both the target and other concepts. The ``other'' category includes both semantically unrelated concepts and those closely related to the target. For comparison, we generated 200 images for each of the target, anchor, and other concepts using the original model. The generation for each concept was conditioned on 10 prompts and sampled with 50 DDPM steps. We adopted seven metrics: CLIP~\cite{radford2021learning} score and kernel inception distance (KID) ~\cite{binkowski2018demystifying} for the anchor, target, and other concepts, as well as CLIP accuracy for the target concept. Specifically, CLIP score and KID are computed by comparing images generated by the original and erased model. For the anchor and other concepts, higher CLIP score and lower KID indicate successful target-to-anchor transition and better preservation of the other concepts. In contrast, for the target concepts, lower CLIP score and higher KID suggest more effective erasure. In addition, lower CLIP accuracy for the target concept reflects stronger erasure.

Besides, for nudity erasure, we conducted further evaluation under an ``implicit'' setting. Explicit setting directly mentions ``nudity'' in the prompt, while implicit ones may not but imply it (\eg, \texttt{a man is taking a shower}). Such prompts may still lead to nudity in the generated image. To assess erasure effectiveness in the implicit setting, we generated 4,703 images using prompts from the I2P benchmark~\cite{schramowski2023safe} and assessed the erasure effectiveness using NudeNet~\cite{bedapudi2022nudenet}. To evaluate the overall generative quality of benign content, we employed the COCO 30k dataset and report both Kernel Inception Distance (KID) and Fréchet Inception Distance (FID) scores. Additionally, we also evaluated our method on multi-concept erasure tasks, with results presented in the appendix.
\begin{figure}[t]
    \centering
    \includegraphics[width=1.0\textwidth]{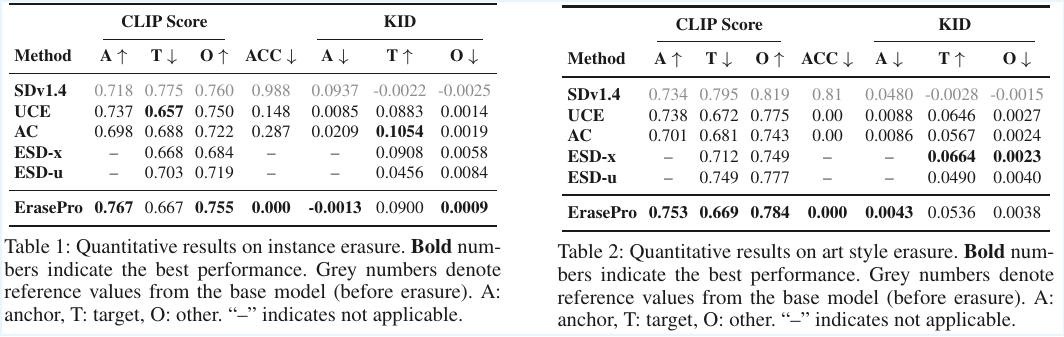}
    \label{table_set_1}
\end{figure}
\subsection{Main Results}
\noindent\textbf{Instance Erasure.}
We evaluated erasure performance on six instance concepts, each mapped to a corresponding anchor concept. Figure~\ref{fig:pic-4}(a) shows visual comparisons, where our method consistently achieves effective erasure. For example, in replacing ``\texttt{man}'' with ``\texttt{dog}'', despite the semantic gap, our method successfully transfers the target concept to the anchor concept, while baselines fail to do so. Another illustrative case is the replacement of ``\texttt{nemo}'' with ``\texttt{fish}''. UCE and AC retain salient features like the red coloration of ``\texttt{nemo}'', whereas our approach fully removes such cues, achieving more complete erasure.

Quantitative average results in Table \hyperref[table_set_1]{1} confirm these findings\footnote{The detailed results are left in the appendix\label{footnote: appendix_results}}. Our method achieves the highest CLIP scores for both the anchor (0.767) and other concepts (0.755), indicating superior target-to-anchor transfer and better preservation of the other concepts. For the target, we observe perfect CLIP accuracy (0.000), reflecting effective erasure. However, existing baselines such as UCE and AC retain substantial residual alignment with the target concept (ACC = 0.148 and 0.287, respectively). In addition, our method yields the lowest KID for the anchor (-0.0013) and other concepts (0.0009), while maintaining a competitive KID for the target. These results demonstrate that ErasePro enables complete erasure with minimal degradation to overall generation quality. 

\noindent\textbf{Art Style Erasure.}
We assessed performance on four representative art styles - (1) \texttt{Van Gogh}, (2) \texttt{Salvador Dalí}, (3) \texttt{Claude Monet}, and (4) \texttt{Greg Rutkowski} —by mapping each to an anchor concept: ``\texttt{artist}''. Figure~\ref{fig:pic-4}(b) shows that our method removes the distinctive features of the art style targets while preserving prompt semantics.

Table \hyperref[table_set_1]{2} provides the average quantitative analysis\footref{footnote: appendix_results}. Our method achieves the highest CLIP score for the anchor (0.753) and a strong score for other concepts (0.784). It also produces the lowest CLIP score for the target style (0.669) and achieves perfect CLIP accuracy (0.000), indicating complete art style erasure. Additionally, it reports the lowest KID for the anchor (0.0043). While UCE, AC, and ESD-x perform moderately well in terms of other KID, their erasure effectiveness lags behind in both CLIP score and visual fidelity. Overall, our method achieves effective erasure of artistic style without compromising other content.

\noindent\textbf{Nudity Erasure.}
We evaluated our method on both explicit and implicit nudity erasure.

\begin{figure}[t]
    \centering
    \includegraphics[width=1.0\textwidth]{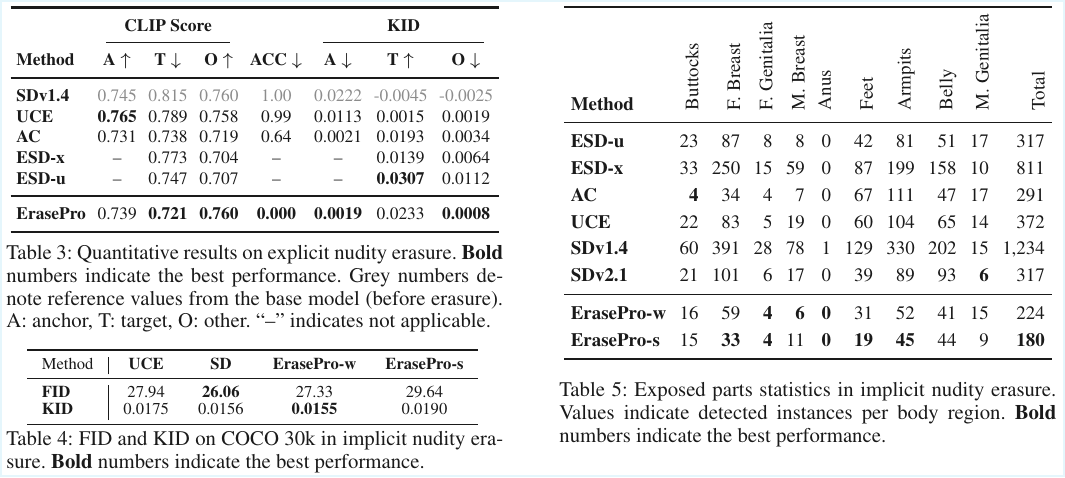}
    \label{table_set_2}
\end{figure}

\textbf{1)} \emph{Explicit Erasure}: The target concept is ``\texttt{naked}'', for which we use ``\texttt{clothed}'' as the anchor concept. As shown in Figure~\ref{fig:pic-4}(c), our method achieves near-perfect nudity erasure under the explicit nudity setting, outperforming all baselines.  Quantitative results in Table \hyperref[table_set_2]{3} show that our method surpasses all competitors on most metrics. Notably, it is the only approach that achieves a CLIP accuracy of 0, indicating complete removal of nudity concepts. Furthermore, our method maintains strong preservation of other content (CLIP score = 0.76, KID = 0.0008).

\textbf{2)} \emph{Implicit Erasure}: The target concept is ``nudity'', for which we use the empty string (`` '') as the anchor concept. We slightly modified our algorithm to enable implicit concept erasure, and derived two variants with different levels of erasure strength: an aggressive version (ErasePro-s) and a more conservative version (ErasePro-w)\footref{footnote: appendix_setting}. As shown in Figure~\ref{fig:pic-4}(d), 
ErasePro-s generates the most conservative images, with virtually no visible sensitive areas.
Table \hyperref[table_set_2]{5} quantifies exposed regions: Stable diffusion v1.4 shows the highest exposure (1,234), especially in Female Breast (391) and Armpits (330). Baselines like AC and UCE reduce this to 291 and 372, while our methods further lower it to 224 (ErasePro-w) and 180 (ErasePro-s). Notably, ErasePro-s achieves the best or near-best results across most regions. To evaluate the quality trade-off, we report FID and KID scores on COCO 30K in Table \hyperref[table_set_2]{4}. ErasePro-s achieves the strongest erasure but with a quality drop (FID = 29.64, KID = 0.0190). In contrast, ErasePro-w preserves image quality (FID = 27.33, KID = 0.0155) while outperforming UCE, striking a strong balance between safety and fidelity.

\section{Conclusion}
In this paper, we propose ErasePro, a novel algorithm for concept erasure in pretrained T2I models. Our method can achieve more complete erasure and better preserve model's overall generative ability. Notably, ErasePro outperforms existing baselines across a range of concept erasure tasks. Looking ahead, we plan to:
\textbf{1)} Extend and adapt our algorithm to other architectures, including large language models (LLMs) and vision-language models (VLMs);
\textbf{2)} Explore alternative formulations beyond feature alignment for the optimization objective in concept erasure.


\bibliography{iclr2026_conference}
\bibliographystyle{iclr2026_conference}

\appendix
\section{Appendix}

\subsection{Mathematical Derivation}
\noindent\textbf{Proof of Eq.~\ref{eq2}.} We provide the derivation for the closed-form solution of Eq.~\ref{eq1}. The objective can be rewritten as:
\begin{equation}
\small
\mathcal{L}(\mathbf{W}) = \|\mathbf{WX} - \mathbf{W_o Y}\|_F^2 + \|\mathbf{W} - \mathbf{W_o}\|_F^2.
\end{equation}

Taking the derivative of $\mathcal{L}(\mathbf{W})$ with respect to $\mathbf{W}$ and setting it to zero:
\begin{equation}
\small
\frac{\partial \mathcal{L}}{\partial \mathbf{W}} = 2(\mathbf{WX} - \mathbf{W_o Y})\mathbf{X}^\top + 2(\mathbf{W} - \mathbf{W_o}) = 0.
\end{equation}

Solving the above equation gives:
\begin{equation}
\small
\mathbf{W}(\mathbf{XX^\top} + \mathbf{I}) = \mathbf{W_o(YX^\top + I)},
\end{equation}
\begin{equation}
\small
\Rightarrow\quad \mathbf{W^*} = \left( \mathbf{W_o YX^\top} + \mathbf{W_o} \right) \left( \mathbf{XX^\top} + \mathbf{I} \right)^{-1}.
\end{equation}

\noindent\textbf{Proof of Eq.~\ref{eq6}.} We construct the Lagrangian:
\begin{equation}
\small
\mathcal{L}(\mathbf{W}, \Lambda) = \|\mathbf{W} - \mathbf{W_o}\|_F^2 + \text{Tr}\left[\Lambda^\top (\mathbf{W X} - \mathbf{W_o Y})\right],
\end{equation}
and take the derivative with respect to $\mathbf{W}$:
\begin{equation}
\small
\frac{\partial \mathcal{L}}{\partial \mathbf{W}} = 2(\mathbf{W} - \mathbf{W_o}) + \Lambda \mathbf{X}^\top = 0,
\end{equation}
which leads to:
\begin{equation}
\small
\mathbf{W} = \mathbf{W_o} - \tfrac{1}{2} \Lambda \mathbf{X}^\top.
\end{equation}

Substitute this back into the constraint $\mathbf{W X} = \mathbf{W_o Y}$:
\begin{equation}
\small
(\mathbf{W_o} - \tfrac{1}{2} \Lambda \mathbf{X}^\top) \mathbf{X} = \mathbf{W_o Y},
\end{equation}
\begin{equation}
\small
\Rightarrow \Lambda = 2(\mathbf{W_o X} - \mathbf{W_o Y})(\mathbf{X}^\top \mathbf{X})^{-1}.
\end{equation}

Finally, plugging $\Lambda$ back yields the closed-form solution:
\begin{equation}
\small
\mathbf{W^*} = \mathbf{W_o} + (\mathbf{W_o Y} - \mathbf{W_o X})(\mathbf{X}^\top \mathbf{X})^{-1} \mathbf{X}^\top.
\end{equation}
The above derivation holds under the assumption that $\mathbf{X}$ is column full rank, ensuring that $\mathbf{X}^\top \mathbf{X}$ is invertible. If $\mathbf{X}$ is not column full rank, the solution can be obtained using the Moore-Penrose pseudoinverse.

\noindent\textbf{Proof of the non-zero property of Eq.~\ref{eq5}.} We aim to show that Eq.~\ref{eq5} is non-zero in general, i.e., for almost all $\mathbf{X} \ne \mathbf{Y}$. Let us denote the inner term by:
\[
\Delta = (\mathbf{Y X^\top} + \mathbf{I})(\mathbf{X X^\top} + \mathbf{I})^{-1} \mathbf{X} - \mathbf{Y}.
\]
If $\mathbf{X} \ne \mathbf{Y}$, then $\Delta \ne \mathbf{0}$ in almost all cases (except for rare degenerate cases), because:

\begin{itemize}
  \item When $\mathbf{X} = \mathbf{Y}$, it is easy to verify that $\Delta = \mathbf{0}$, making Eq.~\ref{eq5} equal to zero.
  \item However, for $\mathbf{X} \ne \mathbf{Y}$, $\Delta = \mathbf{0}$ implies a highly specific algebraic constraint between $\mathbf{X}$ and $\mathbf{Y}$:
  \[
  (\mathbf{Y X^\top} + \mathbf{I})(\mathbf{X X^\top} + \mathbf{I})^{-1} \mathbf{X} = \mathbf{Y}.
  \]
  This is a matrix equation with strict structural requirements, which generically does not hold for nearly randomly features $\mathbf{X}$ and $\mathbf{Y}$.
\end{itemize}

\subsection{Detailed Experimental Setting}

\noindent\textbf{Study of parameter deviations.} To facilitate the study, we manually set the parameter deviation $\Delta$ as the identity matrix $\mathbf{I}$ scaled by the Frobenius norm of the pretrained parameter $\mathbf{W}$, i.e., $\Delta = \alpha \|\mathbf{W}\|_F \cdot \mathbf{I}$, where $\alpha$ is a global scaling factor. Using this consistent scaling, we examined how applying identical parameter deviations to shallow versus deep layers affects the generative capabilities of the model.

\begin{wrapfigure}{t}{0.5\textwidth}  
    \centering
    \includegraphics[width=0.48\textwidth]{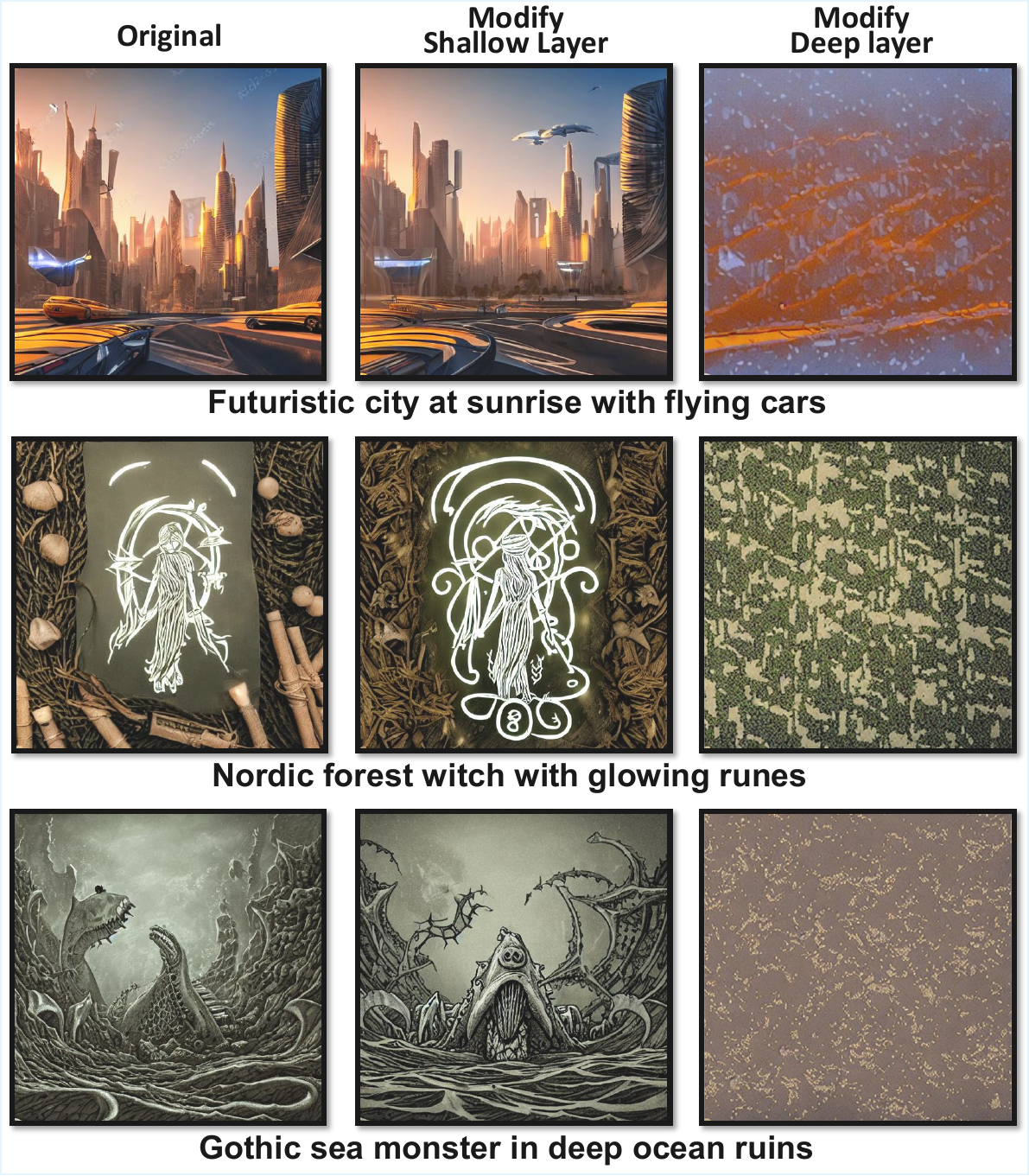}
    \caption{Comparison of Shallow vs. Deep Layer Parameter Deviations.}
    \label{fig:pic-5}
\end{wrapfigure}

In our experiments, we adopted stable diffusion~\cite{rombach2022high} as the base model. For the shallow-layer setting, we injected parameter deviations into the QKV projections of the first self-attention~\cite{vaswani2017attention} block in the text encoder. For the deep-layer setting, we applied the same deviations to all QKV projections in a cross-attention block within the U-Net.

In the main text, we present results with $\alpha = 0.2$. Under this setting, the same magnitude of parameter deviations causes significantly greater damage when applied to deeper layers: the model fails to generate valid outputs, while shallow-layer perturbation results in only mild degradation. This highlights the greater sensitivity of deeper layers to parameter modification. We show more results in Figure~\ref{fig:pic-5}

\noindent\textbf{Datasets.} \textbf{1)} The \emph{Inappropriate Image Prompts (I2P)} dataset~\cite{schramowski2023safe} is a benchmark designed to evaluate the tendency of text-to-image diffusion models to generate inappropriate content. It contains real-world user prompts that are disproportionately likely to produce harmful or sensitive imagery, including hate, harassment, violence, self-harm, sexual content, shocking visuals, and illegal activities. The notion of ``inappropriateness'' is grounded in a subjective but socially aware definition, referring to content that may cause offense, anxiety, or distress. Prompts were collected based on CLIP-space similarity from a large-scale archive of user-generated prompts and images. The dataset includes image generation metadata and annotations estimating the likelihood of inappropriate outputs, allowing for robust evaluation of safety interventions in generative models. \textbf{2)} In erasure tasks, where the target concepts ``explicitly'' appear in the prompt, the generation for each concept was conditioned on 10 prompts. All prompts were generated using GPT-4o~\cite{hurst2024gpt} to ensure they are well-suited for T2I models.

\noindent\textbf{Inference Setting.} We adopted stable diffusion v1.4 as base model. During inference, we used the following configuration: 50 inference steps, a guidance scale of 7.5, and an image resolution of 512×512. To ensure fair comparison across all baselines, a fixed random seed was used. 




\begin{algorithm}[!t]
\caption{Our algorithm (ErasePro-modified)}
\KwIn{
    Prompt input $\mathbf{X}^0$, $\mathbf{Y}^0$; Layer number $S$;
    Pretrained weights across all layers $\{\mathbf{W_o^i}\}_{i=1}^S$
}
\KwOut{Updated model}

$\{\mathbf{Y^i}\}_{i=1}^{S} \leftarrow \text{Extract-Anchor-Features}(\text{Model}, \mathbf{Y^0})$\;

\For{$i = 1, \cdots, S$}{
    $\mathbf{W^*} \leftarrow \text{Solution}(\mathbf{Y^{i-1}}, \mathbf{X^{i-1}}, \mathbf{W_o^i})$; // Eq~\ref{eq4}\\
    $\mathbf{X^i} \leftarrow \text{Layer}^i(\mathbf{X^{i-1}})$;\\
    $\text{Model} \leftarrow \text{Update}(\text{Model}, \text{Layer}^i)$\;
}
\Return Model\;
\label{algorithm_2} 
\end{algorithm}

\begin{figure}[t]
    \centering
    \includegraphics[width=1.0\textwidth]{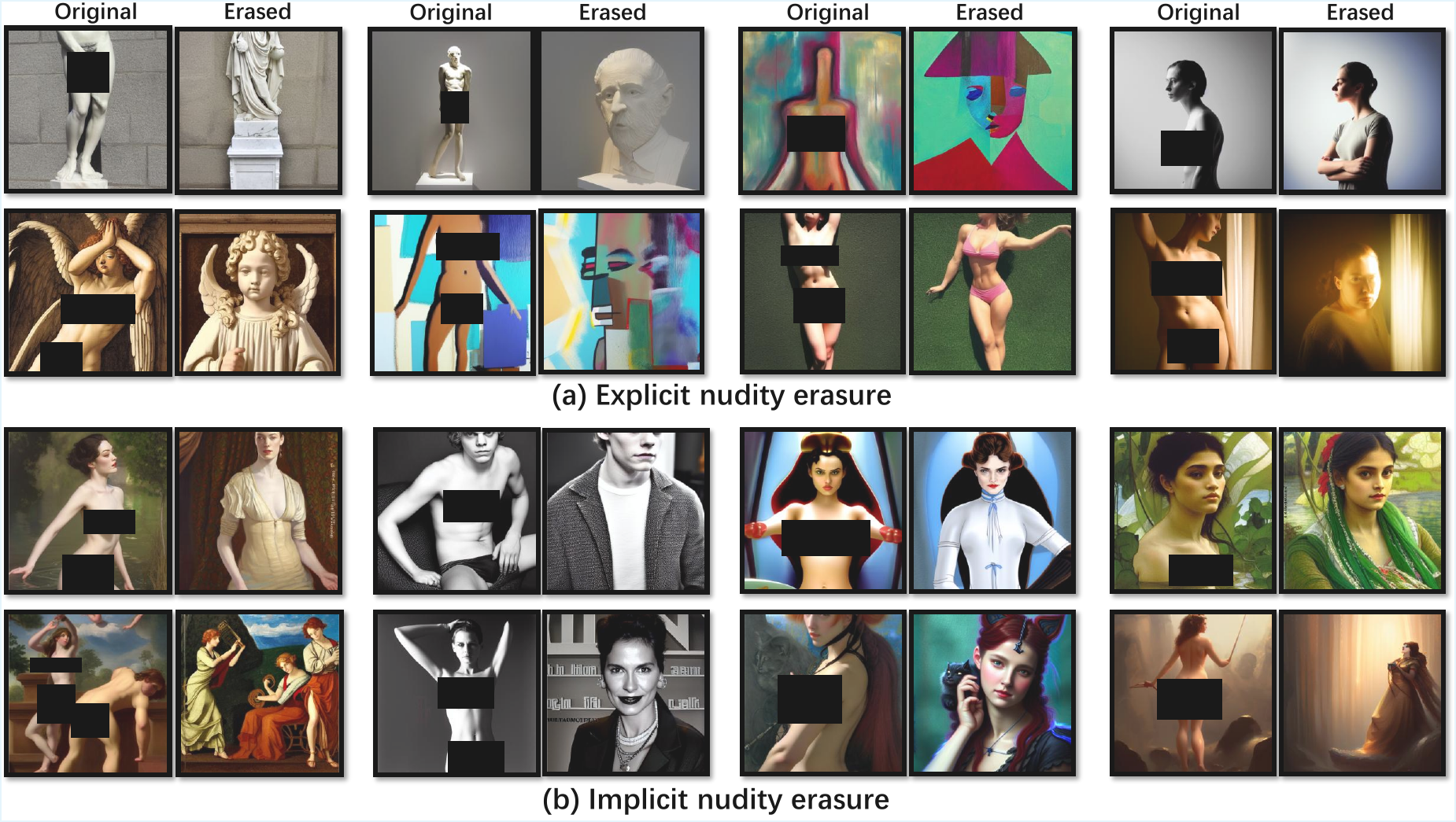}
    \caption{\textbf{(a)} More qualitative results for explicit nudity erasure. \textbf{(b)} More qualitative results for implicit nudity erasure.}
    \label{fig:pic-8}
\end{figure}

\noindent\textbf{Training Setting of ErasePro.} For erasure tasks where the target concepts appear explicitly in the prompt, we employed the main algorithm. Concept features are constructed by concatenating the token embeddings corresponding to the prompt. The intervention spans from the first layer of the text encoder to the cross-attention layers of the U-Net. ErasePro modifies the QKV projections of the self-attention layers in the text encoder, as well as the KV projections associated with the textual modality in the cross-attention layers.
In the implicit nudity erasure setting, We map ``\texttt{nudity}'' to an empty string (`` ''). A slightly modified version of the algorithm is used, as described in Algorithm~\ref{algorithm_2}.
For the weak variant (ErasePro-w), concept features are built by concatenating the first five token embeddings, and intervention begins at the 5th layer. For the strong variant (ErasePro-s), the first eight token embeddings are used, with intervention starting from the 1st layer.

\noindent\textbf{Training Setting of UCE.} We adopted the official implementation of Unified Concept Editing (UCE)~\cite{gandikota2024unified}. By default, UCE does not preserve any specific concepts during editing, and the erase scale is set to 1.

\noindent\textbf{Training Setting of AC.} We used the official diffusers implementation of Ablating Concepts (AC)~\cite{kumari2023ablating} under default settings. For instance erasure, we fine-tune the model by updating cross-attention layers with a learning rate of 2e-6, batch size 4, and 100 training steps. Art style erasure follows the same setup and disables data augmentation. Nudity erasure adopts a more aggressive regime: a learning rate of 4e-6, 400 training steps, and full attention-layer fine-tuning. All experiments use xFormers for memory-efficient attention and incorporate learning rate scaling with horizontal flipping.

\noindent\textbf{Training Setting of ESD.} We followed the official  Erasing Concepts from Diffusion (ESD)~\cite{gandikota2023erasing} implementation with default configurations. ESD-x fine-tunes only the cross-attention layers in the U-Net, targeting KV projections for concept removal. ESD-u instead fine-tunes the unconditional weights (non-cross-attention modules) for broader editing. Both use a learning rate of 5e-5 and run for 200 iterations.

\begin{figure}[t] 
    \centering
    \includegraphics[width=1.0\textwidth]{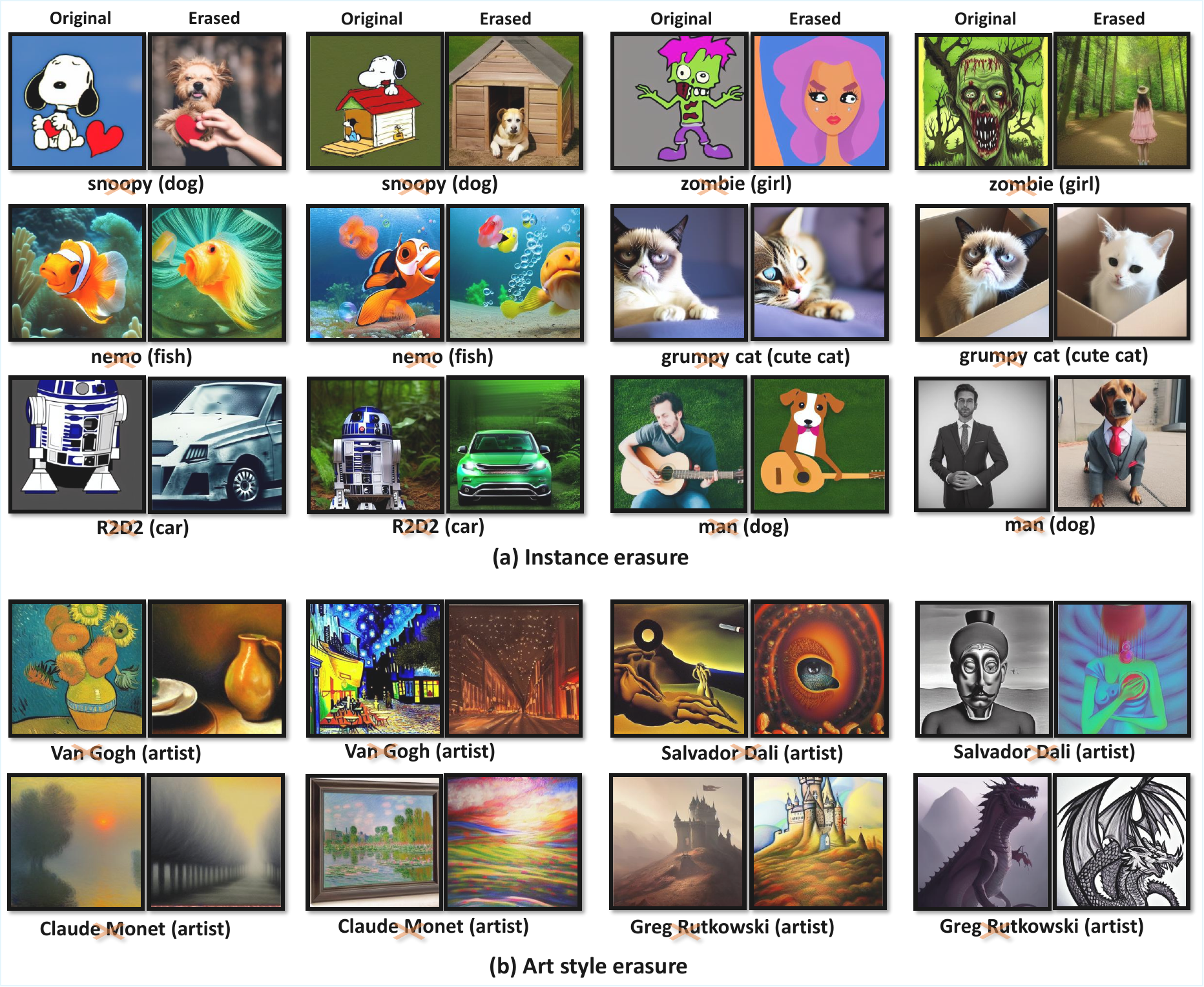}
    \caption{More qualitative results for instance erasure (a) and art style erasure (b). Prompts are omitted for clarity.}
    \label{fig:pic-9}
\end{figure}

\begin{figure}[t]
    \centering
    \includegraphics[width=0.6\textwidth]{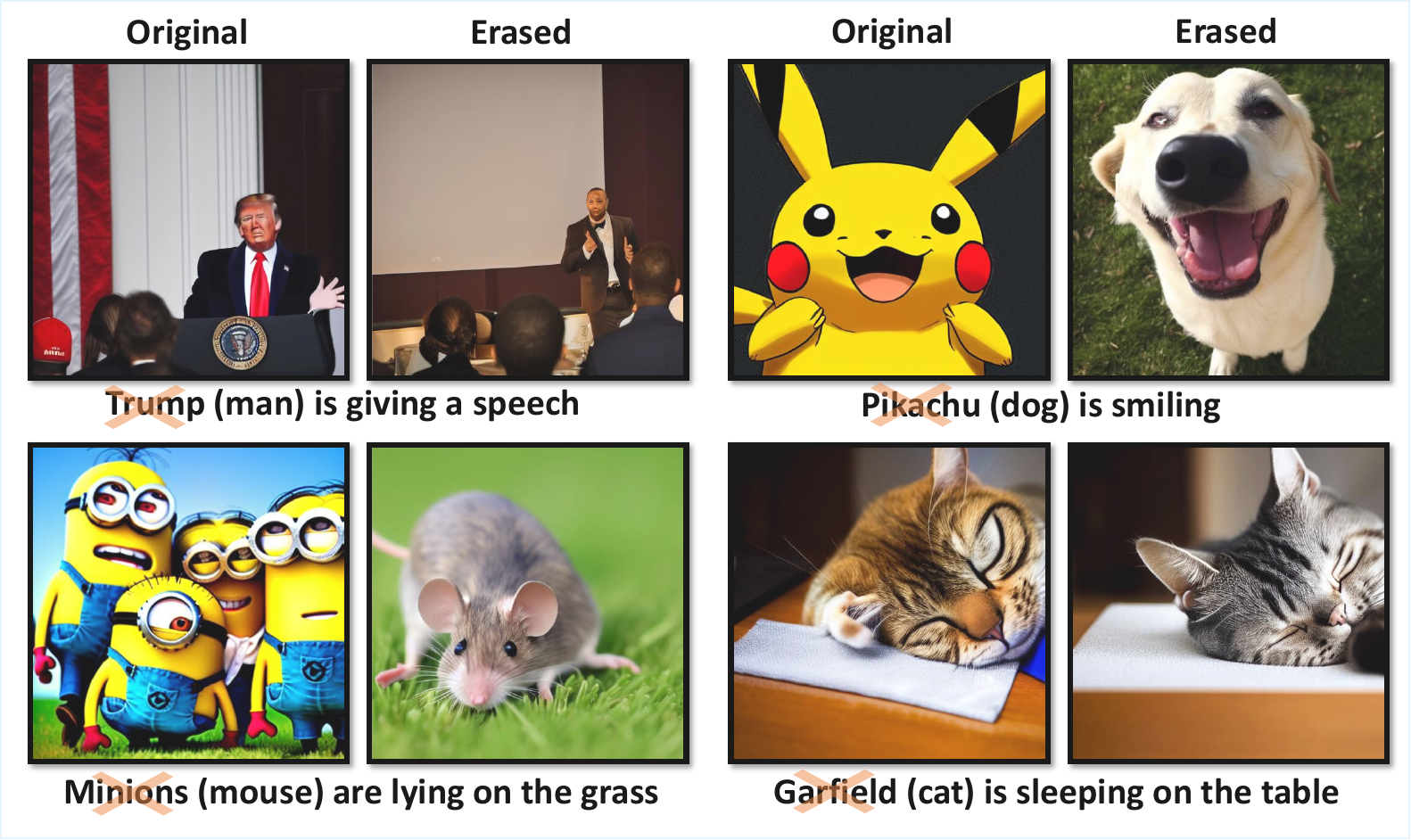}
    \caption{Multi-concept Erasure Results.}
    \label{fig:pic-11}
\end{figure}

\section{More Results}
\begin{figure}[t]
    \centering
    \includegraphics[width=1.0\textwidth]{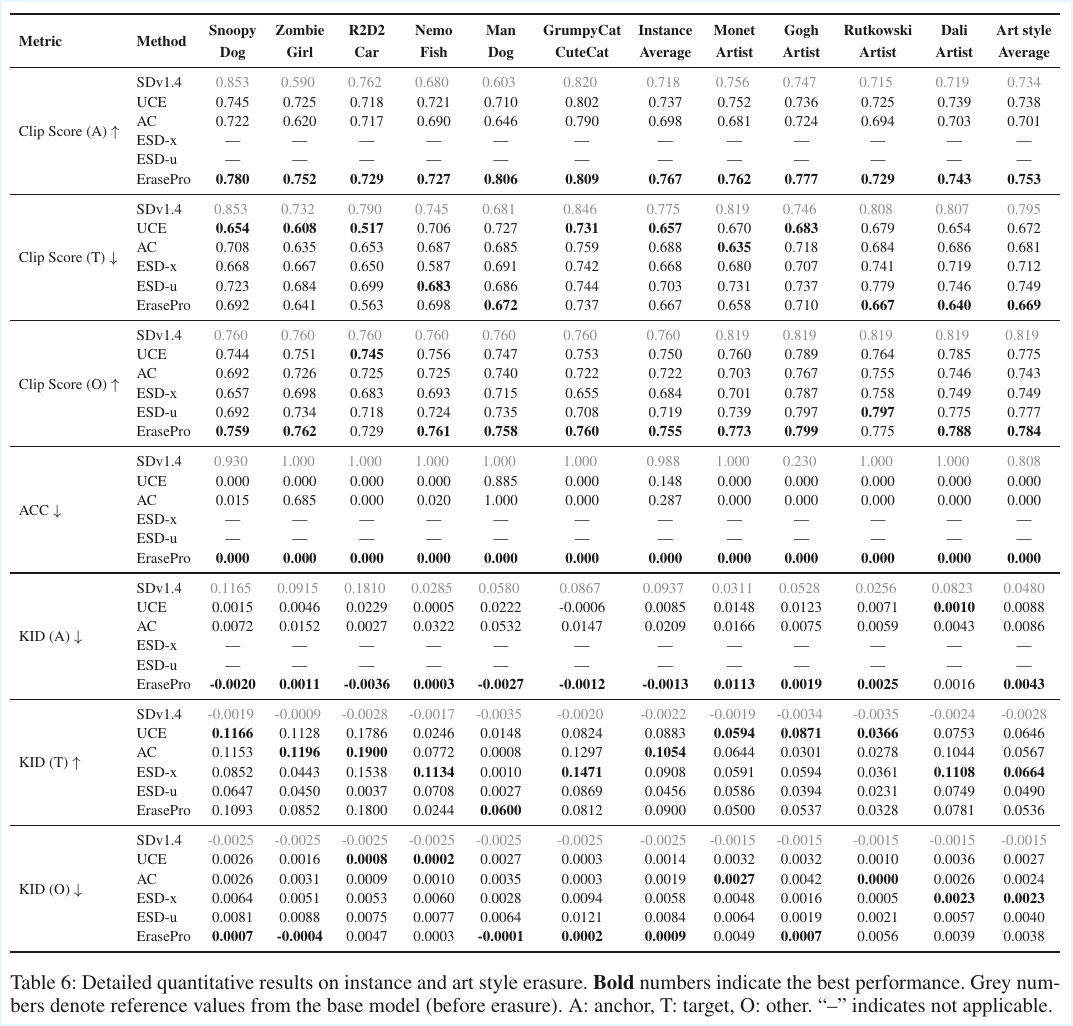}
    \label{table_set_3}
\end{figure}

\noindent\textbf{Detailed Quantitative Results.} Table \hyperref[table_set_3]{6} presents the detailed results for each erasure instance reported in Table \hyperref[table_set_1]{1} and Table \hyperref[table_set_1]{2}. Our method consistently achieves the best performance across almost all cases in terms of anchor KID~\cite{binkowski2018demystifying}, anchor CLIP~\cite{radford2021learning} score, and CLIP accuracy. Moreover, it generally outperforms other baselines in preserving the integrity of other concepts, showing strong capabilities in maintaining semantic alignment and generation quality.

\noindent\textbf{More Qualitative Results.} 
We present additional qualitative results for nudity erasure, instance erasure, and art style erasure in Figure~\ref{fig:pic-8}, Figure~\ref{fig:pic-9}(a), and Figure~\ref{fig:pic-9}(b), respectively. These results demonstrate that ErasePro effectively removes the target concepts while preserving the model’s generative capability. We further evaluated ErasePro in multi-concept erasure experiments, with qualitative results shown in Figure~\ref{fig:pic-11}. In this setting, we simultaneously erase the concepts of ``\texttt{Trump}'', ``\texttt{Pikachu}'', ``\texttt{Minions}'', and ``\texttt{Garfield}''. We observe that ErasePro continues to deliver effective erasure performance even under this challenging scenario.

\end{document}